\documentclass[default, letters]{sn-jnl}

\usepackage{times}
\usepackage{graphicx}
\graphicspath{{./images/}}
\DeclareGraphicsExtensions{.pdf,.png,.jpg}
\usepackage{amssymb, amsmath, mathrsfs}

\usepackage{color}
\usepackage{comment}
\usepackage{subfigure}
\usepackage{tabularx}
\usepackage{multirow}
\usepackage{xspace}
\usepackage{multicol}
\usepackage{xcolor}

\usepackage{wrapfig}

\definecolor{orange}{rgb}{1,0.2,0}
\newcommand{\jc}[1]{{{#1}}}  
\newcommand{\yh}[1]{{{#1}}}

\begin{document}
	
	\title[Automated Catheter Tip Repositioning for Intra-cardiac Echocardiography]{\vspace*{-75pt} \LARGE \bf 
		Automated Catheter Tip Repositioning for Intra-cardiac Echocardiography
	}
	
	\author*[1]{Young-Ho Kim}
	\author[1]{Jarrod Collins}
	\author[2]{Zhongyu Li}
	\author[3]{Ponraj Chinnadurai}
	\author[1]{Ankur Kapoor}
	\author[2,4]{C. Huie Lin}
	\author[1]{Tommaso Mansi}	
	
	\affil[1]{\small Siemens Healthineers, Digital Technology \& Innovation, Princeton, NJ, USA}
	\affil[2]{\small Houston Methodist Research Institute, Houston, TX, USA}
	\affil[3]{\small Siemens Medical Solutions Inc., Advanced Therapies, Malvern, PA, USA}
	\affil[4]{\small Houston Methodist DeBakey Heart \& Vascular Center, Houston, TX, USA \vspace*{-40pt}}
	

\abstract{		
	
{\noindent \bf  Purpose:~}
Intra-Cardiac Echocardiography (ICE) is a powerful imaging modality for guiding cardiac electrophysiology and structural heart interventions. ICE provides real-time observation of anatomy and devices, while enabling direct monitoring of potential complications. In single operator settings, the physician needs to switch back-and-forth between the ICE catheter and therapy device, making continuous ICE support impossible. Two operators setup are therefore sometimes implemented, with the challenge of increase room occupation and cost. \jc{Two operator setups are sometimes implemented, but increase procedural costs and room occupation.}

{\noindent \bf  Methods:~}
ICE catheter robotic control system is developed with automated catheter tip repositioning ({\em i.e.} view recovery) method, which can reproduce important views previously navigated to and saved by the user. The performance of the proposed method is demonstrated and evaluated in a combination of heart phantom and animal experiments. 


{\noindent \bf Results:~}
Automated ICE view recovery achieved catheter tip position accuracy of $2.09~\pm~0.90~mm$ and catheter image orientation accuracy of $3.93~\pm~2.07^\circ$ in animal studies, and $0.67~\pm~0.79~mm$ and $0.37~\pm~0.19^\circ$ in heart phantom studies, respectively. Our proposed method is also successfully used during transeptal puncture in animals without complications, showing the possibility for fluoro-less transeptal puncture with ICE catheter robot. 


{\noindent \bf Conclusion:~}
Robotic ICE imaging has the potential to provide precise and reproducible anatomical views, which can reduce overall execution time, labor burden of procedures, and x-ray usage for a range of cardiac procedures.
}	

	
	
	\maketitle
	
	\vspace*{-25pt}
	\section{Introduction}
	
	Interventional cardiology has expanded its role dramatically in recent years to now encompass treatment of many disease states which were once considered to only have surgical options. This growth has been significantly motivated by the introduction of new treatment devices and advances in intraoperative imaging modalities. Intra-cardiac echocardiography (ICE) has been evolving as a real-time imaging modality for guiding interventional procedures in electrophysiology\,\citep{epstein1998ep, calo2002ep}, congenital\,\citep{rigatelli2005chd, tan2019chd}, and structural heart interventions\,\citep{basman2017shd}, among others. 
	When compared to another more established real-time imaging modality, transesophageal echocardiography (TEE), ICE has improved patient tolerance by not requiring esophageal intubation, requires only local anesthesia with conscious sedation, can be operated by the interventionalist, and does not interfere with fluoroscopic imaging\,\citep{silvestry2009ice}. 
	Real-time ICE imaging has an expanding role in providing uninterrupted guidance for valve replacement interventions\,\citep{bartel2016tavi, saji2016mv}, left atrial appendage closure\,\citep{ren2013laa, matsuo2016laa}, septal defect closure\,\citep{medford2014asd}, and catheter-based ablation for cardiac arrhythmia\,\citep{filgueiras2015abl}. However, with the increased reliance on imaging to perform these complex procedures, there is a high cognitive demand on physicians, who may now be performing both the interventional task and simultaneously acquiring the guiding images. Moreover, they are not always experts in reading ultrasound and navigating these images, which makes ICE handling even more difficult.

	ICE imaging requires substantial training and experience to become comfortable with steering the catheter from within the cardiac anatomy, which hinders its adoption as standard of care\,\citep{bartel2016tavi, tan2019chd}. In practice, the interventionalist needs to continuously manipulate several catheters throughout the procedure, each having different control mechanisms. 
	For example, a typical ablation treatment for cardiac arrhythmia can require tens to hundreds of individual ablations at very specific locations. ICE imaging can be beneficial to monitor for developing complications, target anatomy, facilitate adequate tissue contact, and monitor lesion development during ablations\,\citep{saliba2008abl}. However, this can quickly become a tedious procedure if repositioning are frequently needed.
	Similarly in structural heart procedures, clinicians can manipulate the ICE catheter to localize and measure the area of treatment and then either \textit{park} (\textit{e.g.} to watch for complications) or \textit{retract} the ICE imaging catheter while devices are deployed under fluoroscopic guidance. The ICE catheter is then relocated to visually confirm the placement of therapeutic devices. This manner of repeated manipulation throughout the course of treatment is common for interventions across disciplines but requires intensive coordination, spatial understanding, and manual dexterity that can lead to fatigue in longer or more difficult procedures and imposes a significant learning curve for new users.

	When considering these limitations, it is reasonable that a robotic-assist system that can hold and actively manipulate the ICE catheter, either through operator input or semi-autonomous processes, could ease the workload of the physician during treatment and potentially enable the use of ICE for novel and more complex tasks.
	Several commercial robotic systems for less specific catheter manipulation are currently marketed, including Amigo RCS (Catheter Precision, Inc., Mount Olive, NJ, USA), CorPath GRX (Corindus Inc., Waltham, MA), Magellan, and Sensei (Hansen Medical Inc., Mountain View, CA, USA).
	One commercially-available robotic system for ICE catheter manipulation is the Sterotaxis V-Sono system\,\citep{stereotaxis20}, which controls the ICE catheter robotically, but with reduced degrees-of-freedom. This system provides robotic control of devices via human operators at a remote cockpit based on streamed real-time image (\textit{e.g.} fluoroscopic) feedback. 
	\citet{loschak16ice} have presented a research prototype robotic ICE manipulator and provide a method using electromagnetic (EM) tracking systems to actively maintain focus within the field of view. 
	While some ICE catheters are now manufactured with an EM sensor in the tip for application in ablation procedures, there is an accompanying increase in cost, which is a major consideration for a single-use device. In practice, many commercially available ICE catheters are single use with no position-tracking sensors installed.
	Therefore, the controller for such a robotic-assist system requires an open-loop where spatial feedback are not continuously available. \yh{Accordingly, with this work we introduce a robotic ICE catheter controller that operates without discrete position feedback from added sensors. We believe that this is the first work that demonstrates an automated function for repositioning ICE imaging during a common procedure in animal.}

	\begin{figure}[t!]
	\centering
	\includegraphics[scale= 0.38]{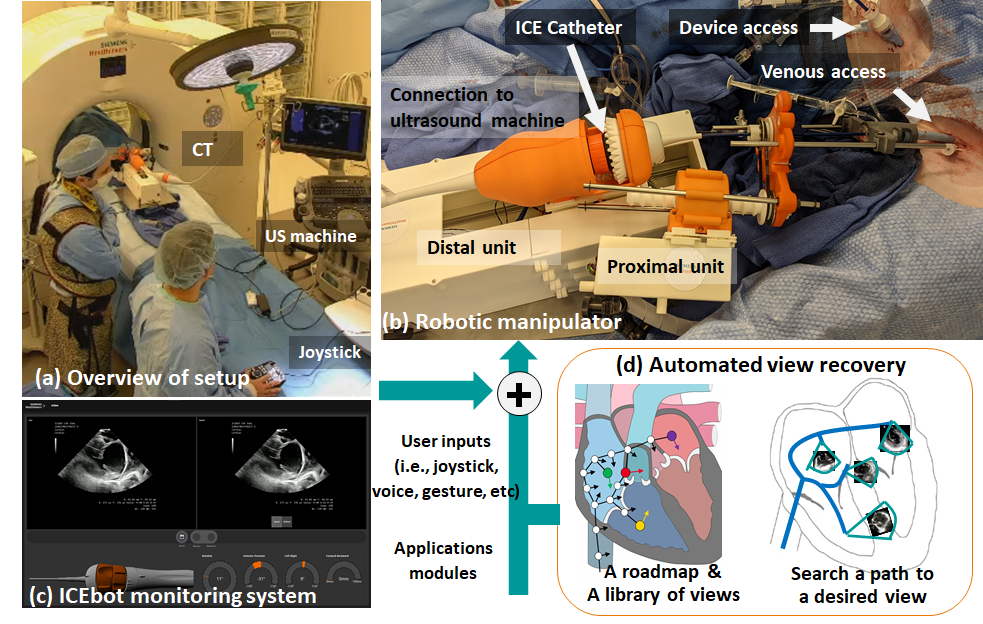}
	\caption{\yh{Overview of the experimental setting, the proposed robotic ICE manipulator, and control scheme: (a)(b) The pig is placed in supine recumbency. The ICE catheter is introduced to the femoral vein at the groin through an introducer sheath by manually, and attached to the ICE manipulator consisted of the proximal/distal units. The joystick input can direct the robotic manipulator. First, the physician-operator manipulates the ICE catheter while surveying images via the ultrasound machine and saving desired views using joystick buttons. Then, whenever the operator request one of the saved views during the procedures, the automated view recovery module can reproduce important views that have been previously saved by the user during the first ICE survey procedure. (c) ICE catheter robot monitoring system shows the list of saved views and lets the operator know the current state of view recovery module ({\em e.g.,} idle, search phase, execution phase, and completed states). (d) Diagram of view recovery process: First, a roadmap that traces motor state configurations is generated while the operator manipulates the ICE catheter by joystic input with each unique configuration represented by a white dot and connected to neighboring configurations by an edge element. Second, a library of desired views is generated by labeling certain states within this roadmap. The colored dots ({\em i.e.,} green, red, yellow, and purple dots) represents the bookmarked or saved state of the robot. Lastly, when the user specify the desired view in the library of views, then the robotic controller can return to any of these views by retracing a path along the roadmap from the current state to the desired state.}
		\vspace*{-10pt}}\label{fig:overview}
	\end{figure}

	

	
	In this paper, we introduce a new robotic controller that can simplify ICE catheter manipulation. 
	Figure\,\ref{fig:overview} presents an overview of the proposed robotic-assist system. 
	As a first step towards automation in clinical ICE workflows, we propose a view recovery method which can autonomously reproduce important views previously navigated to and saved by the user (Section\,\ref{sec:methods}). To achieve this, we implement methods to incrementally generate a topological roadmap of the robotic motor states. When queried, the method provides a path to the specified view by retracing along previous motor configurations and can be reproduced at any time during the procedure. We evaluate our proposed method in a combination of phantom and animal experiments, also during two fluoro-less robot-guided transseptal punctures in animal (Section\,\ref{sec:experiments}).

	\vspace{-5pt}
	\section{Materials and Methods}\label{sec:methods}
	\vspace{-8pt}
	\subsection{ICE catheter robot (ICEbot)}\label{sec:icebot}	
	
\begin{wrapfigure}{r}{4cm}	
		\vspace*{-20pt}
		\includegraphics[width=4cm]{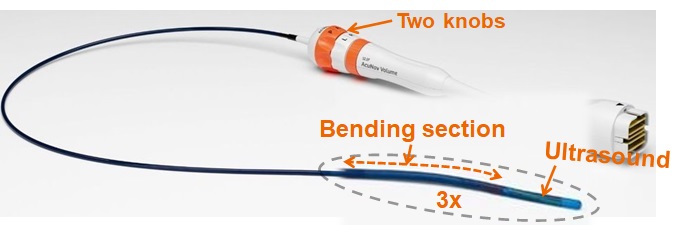}
		\caption{ACUSON AcuNav Volume ICE catheter, Source: Siemens Healthineers }\label{fig:catheter}
		\vspace*{-20pt}
	\end{wrapfigure}

	The ICE catheter (ACUSON AcuNav Volume ICE catheter, Siemens Healthineers) is shown in Figure\,\ref{fig:catheter} with two knobs, bending section, and ultrasound array labeled.  Two pairs of tendon-driven pull mechanisms (one is Anterior-Posterior, nother is Right-Left) consist of a hollow polymer as a sheath and a thread sliding inside the sheath acting as a tendon. Each pair is bound to a common knob, which can pull an individual thread by rotating the knob, allowing one thread to be pulled while the other remains passive. The ultrasound array is located at the tip in the center of the two pairs of tendon mechanisms.
	
	We developed an ICE catheter robotic control system to manipulate the off-the-shelf ICE catheter ({\em i.e.} ACUSON AcuNav Volume ICE catheter family).
	Our robot manipulator consists of two components as shown in Figure\,\,\ref{fig:overview}(b):
	1) A {\it ``proximal"} component holds the catheter shaft, sits directly outside of the introducer sheath, and contributes linear and rotational motion of the catheter. 2) A {\it ``distal"} component holds the catheter handle and controls the two knobs for the bending of the catheter tip, and bulk rotation of the catheter. Moreover, this bulk rotation is synchronized with the \textit{proximal}.
	
	The robot has 4 degrees of freedom. Without loss of generality, we follow the same nomenclature in \,\citep{loschak16ice}: two DOFs for steering the catheter tip in two planes (anterior-posterior knob angle $\phi_1$ and right-left knob angle $\phi_2$) using two knobs on the catheter handle, bulk rotation $\phi_3$, and translation $d_4$ along the major axis of the catheter. We define the robot's configuration, ${\bf q} = (\phi_1,\phi_2,\phi_3,d_4)$ in $\mathbb{R}^4$. 
	\yh{The robot can be controlled with an external joystick (3Dconnexion SpaceMouse pro providing 6DoF analog stick with multiple buttons), which provides a digital input that is directly mapped to the standard knob controls of the catheter or a more intuitive control scheme where the users inputs are directly applied at the catheter tip coordinate frame.}
	
	\vspace{-10pt}
	\subsection{Automated View Recovery: a topological map construction and path planning}\label{sec:view-to-view}
	
	{\small
	\begin{algorithm}[t!]
		\caption{BUILD$\_$ROADMAP$\_$VIEWS~(${\bf q_n}$, $\mathcal{G}$, $\mathcal{V}$, $\epsilon$)}
		\label{algo:graph}
		\begin{algorithmic}[1]
			{
				\State INPUT: the current configuration ${\bf q_n}$, the current roadmap $\mathcal{G}$, the current library of views $\mathcal{V}$, the density parameter $\epsilon$
				\State OUTPUT: $\mathcal{G}$, $\mathcal{V}$
				\State Initialize: ${\bf q_{before}= [~]}$, $\mathcal{G} =[~]$, $\mathcal{V}=[~]$
				\While{ROBOT is OPERATIONAL}			
				\State ${\bf q_n}$ is updated from the current configuration
				\If {$\bf q_{before} \neq  q_n$}
				\For {each $q_i$ $\in$ NEIGHBORHOOD($q_n$,$\mathcal{G}$) }			
				\If {$dist(q_i,q_n) \leq \epsilon$}  
				\If {$q_n$ $\notin$ $\mathcal{G}$} 
				\State 	$\mathcal{G}$.add$\_$vertex($q_n$)\label{step:vertex}
				\EndIf
				\State $\mathcal{G}$.add$\_$edge($q_i$,$q_n$)\label{step:edge}	
				\EndIf	
				\EndFor	
				\If {VIEW$\_$SAVING$\_$FLAG}
				\State $\mathcal{V}$.pushback($q_n$)\label{step:viewsaving}
				\EndIf
				\State ${\bf q_{before} = q_n}$
				\EndIf
				\EndWhile
			}
		\end{algorithmic}
	\end{algorithm}	
}

\yh{Automated view recovery module is used to reproduce important views that have been previously saved by the user, so that the repeated ICE repositioning operations for physicians can be significantly reduced leading the user to focus on the main device handling. The exemplary scenario is following: during an ICE survey for the procedure, the physician-operator continuously manipulates ICE catheter for target anatomies, then the operator picks and saves views using joystick buttons, which is believed as desired views for visually confirming therapeutic devices later. During this process, our proposed module continuously constructs a topological graph ({\it ``a roadmap''}) when the operator manipulates the ICE catheter by joystick input to the robotic manipulator, and generates a library of views (\textit{i.e.} important locations on the \textit{roadmap} that will be revisited later) when the operator clicked the joystick buttons. Once a roadmap and a library of views are constructed by the operator, then the operator can query for a specific view anytime during the procedure. The requested views among a library of views will be given to the controller to search a path from the current pose to the target pose, and then the controller executes motions.}



Figure\,\ref{fig:overview}(c) shows the summary illustration of automated view recovery.
Let $\mathcal{G}({\bf V},{\bf E})$ represent a graph in which {\bf V} denotes the set of configurations ${\bf q_i}$ and {\bf E} is the set of paths $({\bf{q_i}, \bf{q_j}})$. Our proposed automated view recover module is divided into two phases: {\it Construction phase} and {\it Query phase}.

\vspace{5pt}
{\noindent \it Construction phase:}
The library of views and roadmap generation phase: as the robot moves by the operator, the current configuration ${\bf q_n}$ is updated. If ${\bf q_n}$ is not the same as the previous configuration ${\bf q}_{before}$, then the algorithm inserts ${\bf q_n}$ as new vertex of $\mathcal{G}$, 
and connect pairs of ${\bf q_n}$ and existing vertices if the distance is less than a density parameter $\epsilon$ (lines\,\ref{step:vertex}-\,\ref{step:edge} of Algorithm\,\ref{algo:graph}). We apply $\epsilon$ based on Euclidean distance (assuming $1~mm \equiv 1^{\circ}$). If a larger $\epsilon$ (default = 1) were applied, then the search would be faster, however a safety of the path would not be guaranteed, as larger steps along the path could result in collision with anatomy. To validate our approach in this paper, we used $\epsilon = 1$, so generated paths have a millimeter/degree resolution.
Concurrent with roadmap generation, the algorithm constructs a library of views $\mathcal{V}$ when the user saves the view anytime, where $\mathcal{V} = ({{\bf q'_1},...,{\bf q'_m}})$, ${\bf q'}$ is the user saved configuration. $m$ is the number of the user saved views. This step is shown in line\,\ref{step:viewsaving} of Algorithm\,\ref{algo:graph}. 

\vspace{5pt}
{\noindent \it Query phase:}
Given a start configuration ${\bf q_S}$ (the current ${\bf q_n}$) and a goal configuration ${\bf q_G} \in \mathcal{V}$, is given during the procedures. Since each configuration is already in $\mathcal{G}$, we use a discrete ${A^*}$ search algorithm to obtain a sequence of edges that forms a path from ${\bf q_S}$ to ${\bf q_G}$. The more detailed roadmap construction and search algorithms in the graph are in \,\citep{lavalle06planning}.



\vspace{-10pt}
\section{Experiments and Results}\label{sec:experiments}
\vspace{-8pt}
\subsection{Experimental and Validation design}
\subsubsection{Heart phantom study}

\begin{wrapfigure}{r}{3cm}	
	\includegraphics[width=3cm]{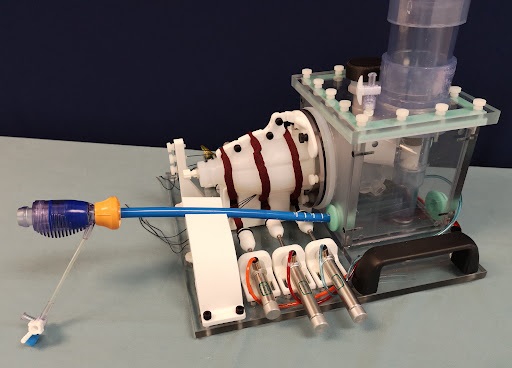}
	\caption{A custom beating heart phantom}\label{fig:heart}
	\vspace*{35pt}
\end{wrapfigure} 

Initial validation of the robotic catheter control during view recovery was performed with the catheter tip inserted into the left ventricle of a custom beating heart phantom (shown in Figure\,\ref{fig:heart}, Archetype Medical) for ICE and TEE valve imaging. An EM sensor (Model 800 sensor, 3D Guidance, Northern Digital Inc.) was attached to the catheter tip to provide real-time tracking as the catheter was manipulated by robotic control.

\yh{This data provides the discrete 3D coordinate and orientation of the catheter tip as it is manipulated when testing the view recovery process. Then, catheter tip position error was calculated as the Euclidean distance between each catheter tip location during view recovery and the respective initial catheter tip location. Similarly, the orientation of the catheter tip is provided by the EM-sensor throughout testing, and the ICE catheter orientation error was calculated.}

\vspace{-10pt}
\subsubsection{Animal study}

Four animal validation experiments were performed at the Houston Methodist Institute for Technology, Innovation $\&$ Education (MITIE, Houston Methodist Hospital) with in vivo study protocol approved by the Institutional Animal Care and Use Committee (IACUC). All testing of the robotic catheter controller was performed under general anesthesia. Vascular access was achieved bilaterally to allow for manipulation of ICE and other device or monitoring catheters. In vivo experimental setup is pictured in Figure\,\ref{fig:overview}. In each experiment, the ICE catheter, with robotic controller pre-attached, was introduced to the venous system through an introducer sheath with balloon seal (DrySeal Flex Introducer Sheath, Gore) before being manually advanced to the junction of the inferior vena cava (IVC) and right atrium (RA) under simultaneous ICE and fluoroscopic guidance.


\yh{In each experiment, initial ultrasound views of the Aortic, Mitral, and Tricuspid valve (i.e. anatomical targets) were manipulated to and saved (i.e. as points-of-interest within the view recovery roadmap) by one physician using joystick input to the robotic controller. Views were then automatically returned to in series several times by the view recovery process. The typical sequence was to consecutively cycle from Tricuspid Valve to Aortic Valve to Mitral Valve in repetition (averaging 4.75 automated positioning events to each valve per experiment).}

\yh{In total, four animal experiments were performed to evaluate view recovery resulting in 45 total view recovery tests. In each view recovery test, following automated robotic motion to the target view, ultrasound and volumetric images were acquired. These data were then compared to measure the reproducibility of robot controlled catheter tip motions.
Dynamic computed tomography (DCT) images were acquired in three of the experiments (30 out of 45) on an Artis zeego (Siemens Healthineers) to provide volumetric ground-truth catheter position within the heart ($200^\circ$ total rotation, 5 second acquisition, 50 frames-per-second). Additionally, traditional gated computed tomography (CT) images (15 out of 45) were acquired in the last experiments on a SOMATOM Force (Siemens Healthineers) to provide volumetric ground-truth catheter position within the heart (0.81 x 0.81 x 0.5~mm voxel spacing). Two-dimensional and volumetric ultrasound imaging was acquired with an ACUSON AcuNav Volume ICE catheter (Siemens Healthineers) \jc{in all experiments}. Ultrasound images were recorded for several heart cycles (typically 3 second acquisition)}.

To validate performance of a tip pose at each view, we compare catheter tip location across multiple robotic positioning events to image targets as the primary spatial validation approach. To measure in vivo, 3D models of the ground-truth observation of the ICE catheter when viewing a target were generated following an intensity-based threshold segmentation of the intra-procedural volumetric imaging ({\em i.e.} DCT or CT) using ITK-SNAP\,\citep{py06nimg}. {Co-registration between corresponding volumetric images from a single experiment was not necessary to align the sequential imaging scans from a given procedure as no large bulk motion ({\em i.e.,} motion of the subject's body relative to the table) was observed between scans.}
Next, the ICE catheter tip was manually labeled from each model and the centroid taken to define the discrete tip location for a given view recovery. For visualization purposes, a curve was then fit to each catheter, constrained to terminate at the predefined catheter tip, by {fitting a fifth-order polynomial}. This process provided a discrete spatial representation of the catheter tip and body. {Finally, catheter tip position error was calculated as the Euclidean distance between each catheter tip location and the respective initial catheter tip location ({\em i.e.,} from the first viewing of the target by the user with joystick controls) for each image target.} 

\yh{Further, to properly validate the change in catheter heading ({\em i.e.} imager orientation), we use a subset of the experimental data from \jc{the final} animal study ({\em i.e.} 15 samples of CT and ultrasound images). This is because DCT was not accurate enough to distinguish \jc{the imager orientation relative to} anatomical targets. Thus, this validation of orientation is accomplished \jc{in the final animal study} by co-registering the volumetric ultrasound data with the corresponding CT images for a given view recovery.} To achieve, corresponding anatomy were labeled in both the volumetric ultrasound and CT data ({\em e.g.,} aortic valve, mitral valve, and tricuspid valve) and the centroid taken. The ultrasound transducer location ({\em i.e.} from the ICE catheter) is also known in each image space. 
\yh{A corresponding point-based registration was then performed with the transducer and anatomies equally weighted. Corresponding points were selected and labeled ({\em i.e.,} fiducials) within each image-space ({\em e.g., } ICE catheter tip, tricuspid valve centroid, etc). Co-registration was achieved by performing a rigid-body, corresponding point-based alignment by the minimization of the sum of squares of fiducial registration errors\,\citep{peter66procrustes,fitzpatrick98registration}.} Since the ultrasound orientation is trivial to determine in the ultrasound image space, it can then be transformed to the CT image space. Finally, the transducer orientation error, $\alpha$, was calculated between each catheter orientation and the respective initial catheter orientation for each image target by equation\,\eqref{eq:orientation}. 
{\small
\begin{equation}
cos(\alpha) = \frac{\hat{a}\cdot\hat{b}}{\|\hat{a}\|\cdot\|\hat{b}\|}\label{eq:orientation}.
\end{equation}
}
For the volumetric ultrasound data, each anatomy was manually labeled 5 times per acquisition and the average taken ({\em i.e.} resulting in 25 samples per target valve, 75 total samples). Intra-observer variability was measured as $1.64 \pm 1.15~mm$ in labeling anatomy which translates to $1.95 \pm 1.55^\circ$ error in calculation of the orientation.

\vspace{-10pt}
\subsection{Results}\label{sec:results}\vspace{-5pt}
\subsubsection{Heart phantom study results}

\yh{Four unique catheter positions were manipulated to within the beating heart phantom by joystick input to the robotic controller and saved for automated recovery. The set of target positions encompassed manipulation of each of the 4-DoF of catheter motion. The controller was then tasked with cycling the ICE catheter tip between each target position in series a total of 16 times. 
Accuracy of the robotic controller was measured by an EM sensor attached at the catheter tip. The set of target positions encompassed manipulation of all 4 DOF of catheter motion. Under these conditions, the robotic controller maneuvered the catheter tip to the target position with an average position error of $0.67 \pm 0.79~mm$ and rotational error of $0.37 \pm 0.19^{\circ}$.}

\begin{figure}[t!]
	\centering
	\includegraphics[scale= 0.5]{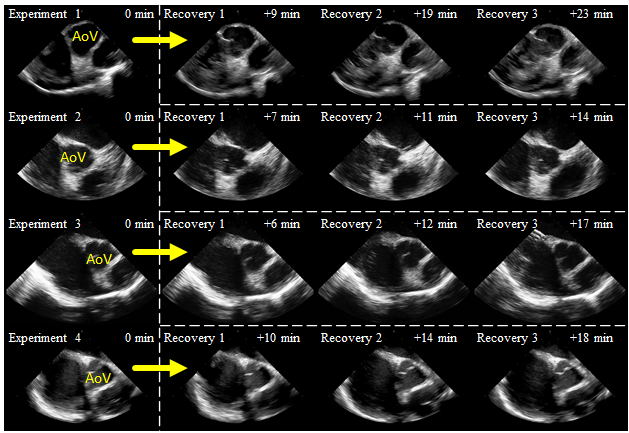}
	\caption{Longitudinal aortic valve images from each of the four experiments reported in this study. Each row represents a separate experiment. The left-most column represents the initial viewing of the valve. Moving right-ward each further column represents subsequent returns to the view at the annotated times.
	}\label{fig:viewrecovery_results}
	\vspace*{-15pt}
\end{figure}

\subsubsection{Animal study results}

\begin{figure}[t]
	\centering
	\includegraphics[scale= 0.6]{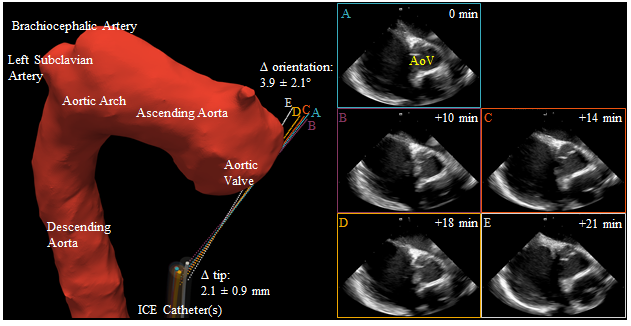}
	\caption{ (Left) Three-dimensional representation of aortic anatomy (red) and each of 5 ICE catheters as they image the aortic valve when testing the view recovery process in vivo. ICE catheter orientation is represented by the corresponding dotted lines. (Right) Ultrasound images corresponding to each ICE catheter. Where A represents the initial viewing of the Aortic Valve and B-D represent images after applying view recovery to return from within the heart to this view. The time of each image relative to the initial viewing of the target (A) are noted.		
	\vspace{-10pt}}\label{fig:viewrecovery_results2}
\end{figure}

Catheter tip localization and catheter orientation error are presented in Table\,\ref{tab:table1} for each imaging target. Across all experimental data and image-targets, average catheter tip localization error relative to the respective initial target viewing was $2.09\pm 0.90~mm$ (n = 45). Further, average catheter orientation error relative to the initial target viewing was $3.93 \pm 2.07^\circ$ (n = 15). A subset of longitudinal aortic valve images from each experiment are presented in Figure\,\ref{fig:viewrecovery_results} for qualitative evaluation – with each row representing a separate animal experiment, the first column presenting the initial viewing of the aortic valve, and subsequent columns presenting automated recoveries of the aortic valve at the given times. Figure\,\ref{fig:viewrecovery_results2} displays the ground-truth positioning of each ICE catheter relative to the aorta ({\em i.e.} from contrast-enhanced gated CT). Corresponding ICE catheter orientations are displayed (i.e. from co-registering volumetric ultrasound and CT) as well as the resulting ultrasound images.


\begin{table*}[t]
	\centering{	\small	
		\begin{tabular}{|l|c|c|}
			\hline
			&		 Catheter tip position error [mm] & Imager orientation error [$^\circ$] \\ 		
			\hline
			Aortic Valve        & $2.19\pm0.91$ & $3.08\pm2.49$   \\
			Mitral Valve        & $1.71\pm0.74$ & $4.87\pm2.09$    \\
			Tricuspid Valve     & $2.38\pm0.96$ & $3.85\pm1.70$    \\ \hline
	\end{tabular}}
	\vspace{3pt}
	\caption{Catheter tip position (45 samples) and orientation error (15 samples): Mean and standard deviation are across four experiments. All are measured relative to initial viewing}\label{tab:table1}
	\vspace*{-15pt}
\end{table*}

%

\begin{figure}[t]
	\centering
	\includegraphics[scale= 0.5]{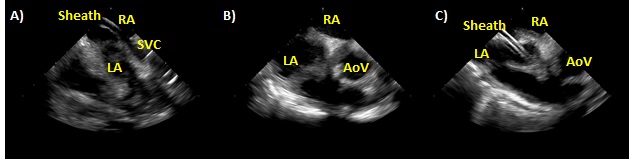}
	\caption{Application of transeptal puncture procedure: (A) Saved view 1, superior vena cava view prior to puncture. Sheath is visible in right atrium abutting the septal wall. (B) Saved view 2, septal view prior to puncture. (C) Saved view 2, septal view post puncture. Sheath has been placed through the puncture across the septal wall. autonomous view recovery was used to manipulate ICE catheter between the two pre-saved views as needed.
		\label{fig:transeptal}}
\end{figure}

Transeptal puncture is routinely performed in electrophysiology and structural heart procedures. With the initial development of ICE imaging in the previous two decades, ICE-guided (i.e. zero-fluoroscopy) transeptal procedures were hypothesized. However, to date, fluoroscopy remains the imaging modality of choice for guiding transseptal therapies. 
Robotic ICE guidance using automated view recovery was performend for a transeptal puncture in two animals. The physician-operator began by pre-saving two ICE views from the RA relevant to the procedure (Figure\,\ref{fig:transeptal}~ A and B). The puncture was achieved with a Versacross RF (Baylis Medical). The procedure was performed entirely with robotic ICE guidance (i.e. no fluoroscopy or other real-time imaging). After pre-saving of the two guiding views, the operator did not need to maneuver the ICE via joystick or manually for the remainder of the procedure. Automated view recovery was utilized to manipulate between the two pre-saved views as needed (with 5 transitions from view-to-view in each experiment). Real-time ICE view swapping was achieved by button-press when desired. Following the procedure, the puncture was verified first by bubble study and then by placing a sheath across the septal wall and acquiring a CT with the left heart contrast-enhanced via the septal sheath. The detailed results are recently presented in clinical society\,\citep{zhongyu21ice}, showing that transeptal puncture was successfully achieved without complications, and automatic view recovery of ICE robotic manipulator might be a useful tool for ICE clinical applications.




\section{Discussion}\label{sec:discussion}

\yh{In practice, standardized anatomical imaging views are emerging as indications for ICE continue to expand across disciplines. \citet{enriquez2018use} detail standard imaging views for a variety of structural and electrophysiology procedures. Similarly, early identification and unobstructed monitoring of potentially life-threatening complications is one of the most valuable functions that ICE provides \,\citep{filgueiras2015abl} and is leading to its adoption as a primary imaging modality for certain procedures, especially those involving transseptal puncture for left heart catheterization \,\citep{silvestry2009ice}.}

Relative to typical right atrial measurements from echocardiography ($3.6 \pm 0.1$ cm and $4.2 \pm 0.1$ cm for short- and long axis respectively\,\citep{bommer1979determination}), the level of catheter tip placement error ({\em i.e.} $2.1 \pm 0.9~mm$) and imager orientation error ({\em i.e.} $3.9 \pm 2.1^\circ$) observed in this study demonstrate high accuracy for automated view recovery. These results are further corroborated by the phantom catheter tip localization results acquired by EM sensing (i.e. $0.67~mm$ and $0.37^\circ$). However, we must note that the measurements taken in phantom are below the reported accuracy for localization ({\em i.e.} $1.4~mm$) and orientation ({\em i.e.} $0.5^\circ$) and should be considered as such. Comparing these in vivo and phantom results demonstrates a compounding of error sources when moving between experimental settings which include: mechanical coupling of the catheter, robot, and sheath; interactions between the catheter and vascular anatomy; and physiological sources such as cardiac or respiratory motion. Altogether, we believe that these results support our conclusion that the robotic controller and automated view recovery process introduced in this study can accurately position the ICE catheter within the RA and reproducibly image cardiac structures of clinical relevance.

\yh{Safety is a major concern when considering any degree of robotic autonomy in human interfacing applications. 
As described, the automated view recovery method is not autonomously finding views in an active manner, but rather relying on the user's initialization of each desired view and their tracing of possible paths (\textit{i.e.} roadmap $\mathcal{G}$). Thus a safe navigation highly relies on the user's initial controls.
Then, how can we generate a reliable tracing during the manual operation to gather tracing and a library of views? First, the user may cause the catheter to contact critical anatomies (\textit{e.g.} septal wall, valves). However this might be acceptable because the ICE catheter's end portion and tip are designed to be yielding so that the catheter will not puncture or damage tissue. Second, \jc{physicians continuously watch} the ICE images while they manipulate ICE catheters \jc{to survey anatomy.} \jc{Therefore, real-time ICE image feedback helps to guide the user during manipulation, creating a safe path.} In addition, the critical safety issues can be handled by robotic manipulation too. For example, risky actions might be easily detected and banned by force controls. Once we have a reliable tracing and the library of views, then with our presented semi-autonomous functionality we can recover the user's previous navigation including uncertainties ({\em i.e.} interaction with vessel and heart beating). Finally, our spatial- and image-based performance results support that the robotic controller can maintain a certain accuracy accordance with the expert user's trace.}

\yh{In each animal study, we saved around 8000 states in the graph, which is about 40 min to 1 hour worth of operations. There are not many crossover nodes in constructed graph, thus the path search time using $A^*$ algorithms was less than one second in our laptop setting (Alienware m15, 9th Intel i7, 32GB memory).
If the size of the graph increased, this might not be in real-time operation. These optimization of graph search is not covered in this paper. Instead, we focused on the feasibility studies for automated view recovery.}




\section{Conclusion}\label{sec:conclusion}


Intra-cardiac Echocardiography (ICE) is a powerful imaging modality for guiding electrophysiology and structural heart interventions. ICE provides real-time observation of anatomy, catheters, and emergent complications. However, this increased reliance on intraprocedural imaging creates a high cognitive demand on physicians who can often serve as interventionalist and imager. Therefore, we developed automated view recovery for ICE catheter robot to assist physicians with imaging and serve as a platform for developing processes for procedural automation. The automated view recovery method allows physicians to save views during intervention and automatically return with the push of a button.
The proposed method was validated by repeated catheter positioning in cardiac phantom with EM-sensor and animal experiments with image-based analysis. 
Results support that a robotic manipulator for ICE can provide an efficient and reproducible tool, potentially reducing execution time and promoting greater utilization of ICE imaging. This work contributes to advancement in the application of ICE in electrophysiology and structural heart procedures.


	
	\section*{Disclaimer}
	{	
		The concepts and information presented in this paper are based on research results that are not commercially available. Future availability cannot be guaranteed.		
	}

	{
		\footnotesize
		\bibliography{references_view}	

\begin{thebibliography}{22}
\providecommand{\natexlab}[1]{#1}
\providecommand{\url}[1]{#1}
\csname url@samestyle\endcsname
\providecommand{\newblock}{\relax}
\providecommand{\bibinfo}[2]{#2}
\providecommand{\BIBentrySTDinterwordspacing}{\spaceskip=0pt\relax}
\providecommand{\BIBentryALTinterwordstretchfactor}{4}
\providecommand{\BIBentryALTinterwordspacing}{\spaceskip=\fontdimen2\font plus
\BIBentryALTinterwordstretchfactor\fontdimen3\font minus
  \fontdimen4\font\relax}
\providecommand{\BIBforeignlanguage}[2]{{%
\expandafter\ifx\csname l@#1\endcsname\relax
\typeout{** WARNING: IEEEtranN.bst: No hyphenation pattern has been}%
\typeout{** loaded for the language `#1'. Using the pattern for}%
\typeout{** the default language instead.}%
\else
\language=\csname l@#1\endcsname
\fi
#2}}
\providecommand{\BIBdecl}{\relax}
\BIBdecl

\bibitem[Epstein et~al.(1998)Epstein, Smith, and TenHoff]{epstein1998ep}
L.~Epstein, T.~Smith, and H.~TenHoff, ``Nonfluoroscopic transseptal
  catheterization: safety and efficacy of intracardiac echocardiographic
  guidance,'' \emph{Journal of Cardiovascular Electrophysiology}, vol.~9,
  no.~6, pp. 625--630, 1998.

\bibitem[Calo et~al.(2002)Calo, Lamberti, Loricchio, D'Alto, Castro, Boggi, and
  et~al.]{calo2002ep}
L.~Calo, F.~Lamberti, M.~Loricchio, M.~D'Alto, A.~Castro, A.~Boggi, and et~al.,
  ``Intracardiac echocardiography: from electroanatomic correlation to clinical
  application in interventional electrophysiology,'' \emph{Italian Heart
  Journal}, vol.~3, no.~7, pp. 387--398, 2002.

\bibitem[Rigatelli(2005)]{rigatelli2005chd}
G.~Rigatelli, ``Expanding the use of intracardiac echocardiography in
  congenital heart disease catheter-based interventions,'' \emph{Journal of the
  American Society of Echocardiography}, vol.~18, pp. 1230--1231, 2005.

\bibitem[Tan and Aboulhosn(2019)]{tan2019chd}
W.~Tan and J.~Aboulhosn, ``Echocardiographic guidance of interventions in
  adults with congenital heart defects,'' \emph{Cardiovascular Diagnosis and
  Therapy}, vol.~9, no.~2, pp. S346--S359, 2019.

\bibitem[Basman et~al.(2017)Basman, Parmar, and Kronzon]{basman2017shd}
C.~Basman, Y.~Parmar, and I.~Kronzon, ``Intracardiac echocardiography for
  structural heart and electrophysiological interventions,'' \emph{Current
  Cardiology Reports}, vol.~19, no.~10, p. 102, 2017.

\bibitem[Silvestry et~al.(2009)Silvestry, Kerber, Brook, Carroll, Eberman,
  Goldstein, and et~al.]{silvestry2009ice}
F.~Silvestry, R.~Kerber, M.~Brook, J.~Carroll, K.~Eberman, S.~Goldstein, and
  et~al., ``Echocardiography-guided interventions,'' \emph{Journal of the
  American Society of Echocardiography}, vol.~22, no.~3, pp. 213--231, 2009.

\bibitem[Bartel et~al.(2016)Bartel, Edris, Velik-Salchner, and
  Muller]{bartel2016tavi}
T.~Bartel, A.~Edris, C.~Velik-Salchner, and S.~Muller, ``Intracardiac
  echocardiography for guidance of transcatheter aortic valve implantation
  under monitored sedation: a solution to a dilemma?'' \emph{European Heart
  Journal of Cardiovascular Imaging}, vol.~17, no.~1, pp. 1--8, 2016.

\bibitem[Saji et~al.(2016)Saji, Rossi, Ailawadi, Dent, Ragosta, and
  Lim]{saji2016mv}
M.~Saji, A.~Rossi, G.~Ailawadi, J.~Dent, M.~Ragosta, and D.~Lim, ``Adjunctive
  intracardiac echocardiography imaging from the left ventricle to guide
  percutaneous mitral valve repair with the mitraclip in patients with failed
  prior surgical rings,'' \emph{Catheterization and Cardiovascular
  Interventions}, vol.~87, no.~2, pp. e75--e82, 2016.

\bibitem[Ren et~al.(2013)Ren, Marchlinksy, Supple, Hutchinson, Garcia, Riley,
  and et~al.]{ren2013laa}
J.~Ren, F.~Marchlinksy, G.~Supple, M.~Hutchinson, F.~Garcia, M.~Riley, and
  et~al., ``Intracardiac echocardiographic diagnosis of thrombus formation in
  the left atrial appendage: a complementary role to transesophageal
  echocardiography,'' \emph{Echocardiography}, vol.~30, pp. 72--80, 2013.

\bibitem[Matsuo et~al.(2016)Matsuo, Neuzil, Petru, Chovanec, Janotka, Choudry,
  and et~al.]{matsuo2016laa}
Y.~Matsuo, P.~Neuzil, J.~Petru, M.~Chovanec, M.~Janotka, S.~Choudry, and
  et~al., ``Left atrial appendage closure under intracardiac echocardiographic
  guidance: feasibility and comparison with transesophageal echocardiography,''
  \emph{Journal of the American Heart Association}, vol.~5, no.~10, p.~4, 2016.

\bibitem[Medford et~al.(2014)Medford, Taggart, Cabalka, Cetta, Reeder, Hagler,
  and et~al.]{medford2014asd}
B.~Medford, N.~Taggart, A.~Cabalka, F.~Cetta, G.~Reeder, D.~Hagler, and et~al.,
  ``Intracardiac echocardiography during atrial septal defect and patent
  foramen ovale device closure in pediatric and adolescent patients,''
  \emph{Journal of the American Society of Echocardiography}, vol.~27, no.~9,
  pp. 984--990, 2014.

\bibitem[Filgueiras-Rama et~al.(2015)Filgueiras-Rama, de~Torres-Alba,
  Castrejon-Castrejon, Estrada, Figueroa, Salvador-Montanes, and
  et~al.]{filgueiras2015abl}
D.~Filgueiras-Rama, F.~de~Torres-Alba, S.~Castrejon-Castrejon, A.~Estrada,
  J.~Figueroa, O.~Salvador-Montanes, and et~al., ``Utility of intracardiac
  echocardiography for catheter ablation of complex cardiac arrhythmias in a
  medium-volume training center,'' \emph{Echocardiography}, vol.~32, no.~4, pp.
  660--670, 2015.

\bibitem[Saliba and Thomas(2008)]{saliba2008abl}
W.~Saliba and J.~Thomas, ``Intracardiac echocardiography during catheter
  ablation of atrial fibrillation,'' \emph{Europace}, vol.~10, no.~3, pp.
  42--47, 2008.

\bibitem[Stereotaxis(2020)]{stereotaxis20}
Stereotaxis, ``Stereotaxis v-drive robotic navigation system,'' Feb. 2020,
  {http://www.stereotaxis.com/products/vdrive}.

\bibitem[{Loschak} et~al.(2017){Loschak}, {Brattain}, and {Howe}]{loschak16ice}
P.~M. {Loschak}, L.~J. {Brattain}, and R.~D. {Howe}, ``Algorithms for
  automatically pointing ultrasound imaging catheters,'' \emph{IEEE
  Transactions on Robotics}, vol.~33, no.~1, pp. 81--91, Feb 2017.

\bibitem[LaValle(2006)]{lavalle06planning}
S.~M. LaValle, \emph{Planning Algorithms}.\hskip 1em plus 0.5em minus
  0.4em\relax New York, NY, USA: Cambridge University Press, 2006.

\bibitem[Yushkevich et~al.(2006)Yushkevich, Piven, Cody~Hazlett, Gimpel~Smith,
  Ho, Gee, and Gerig]{py06nimg}
P.~A. Yushkevich, J.~Piven, H.~Cody~Hazlett, R.~Gimpel~Smith, S.~Ho, J.~C. Gee,
  and G.~Gerig, ``User-guided {3D} active contour segmentation of anatomical
  structures: Significantly improved efficiency and reliability,''
  \emph{Neuroimage}, vol.~31, no.~3, pp. 1116--1128, 2006.

\bibitem[Schönemann(1966)]{peter66procrustes}
P.~H. Schönemann, ``A generalized solution of the orthogonal procrustes
  problem,'' \emph{Psychometrika}, vol.~31, pp. 1--10, 1966.

\bibitem[Fitzpatrick et~al.(1998)Fitzpatrick, West, and
  Maurer]{fitzpatrick98registration}
J.~Fitzpatrick, J.~West, and C.~Maurer, ``Predicting error in rigid-body
  point-based registration,'' \emph{IEEE Transactions on Medical Imaging},
  vol.~17, no.~5, pp. 694--702, 1998.

\bibitem[Li et~al.(2021)Li, Collins, Kim, Chinnadurai, Mansi, and
  Lin]{zhongyu21ice}
Z.~Li, J.~Collins, Y.-H. Kim, P.~Chinnadurai, T.~Mansi, and C.~H. Lin,
  ``Zero-fluoroscopy transseptal puncture guided by intelligent intracardiac
  echocardiography robotics,'' \emph{Journal of the American College of
  Cardiology}, vol.~77, no. 18\_Supplement\_1, pp. 970--970, 2021.

\bibitem[Enriquez et~al.(2018)Enriquez, Saenz, Rosso, Silvestry, Callans,
  Marchlinski, and Garcia]{enriquez2018use}
A.~Enriquez, L.~C. Saenz, R.~Rosso, F.~E. Silvestry, D.~Callans, F.~E.
  Marchlinski, and F.~Garcia, ``Use of intracardiac echocardiography in
  interventional cardiology: working with the anatomy rather than fighting
  it,'' \emph{Circulation}, vol. 137, no.~21, pp. 2278--2294, 2018.

\bibitem[Bommer et~al.(1979)Bommer, Weinert, Neumann, Neef, Mason, and
  DeMaria]{bommer1979determination}
W.~Bommer, L.~Weinert, A.~Neumann, J.~Neef, D.~T. Mason, and A.~DeMaria,
  ``Determination of right atrial and right ventricular size by two-dimensional
  echocardiography,'' \emph{Circulation}, vol.~60, no.~1, pp. 91--100, 1979.

\end{thebibliography}
	}

\end{document}